\documentclass[letterpaper, 10 pt, conference]{ieeeconf}  

\IEEEoverridecommandlockouts                              

\usepackage{graphicx}
\usepackage{caption}
\usepackage{amsmath} 
\usepackage{amssymb}
\usepackage{amsthm}
\usepackage{easytable}
\usepackage{mathtools}
\usepackage{mathrsfs}
\usepackage[mathscr]{euscript}
\usepackage{times}
\usepackage{autobreak}
\usepackage[ruled,vlined,linesnumbered]{algorithm2e}
\usepackage{algpseudocode}

\usepackage{breqn}
\usepackage{color}
\usepackage{comment}
\usepackage{subcaption}
\usepackage{booktabs}
\usepackage[usenames,dvipsnames,svgnames,table, xcdraw]{xcolor}
\usepackage[colorlinks=true,pdfpagemode=UseNone,citecolor=black,linkcolor=black,urlcolor=BrickRed]{hyperref}
\usepackage{diagbox}
\usepackage{amsfonts}
\usepackage[utf8]{inputenc}
\usepackage[T1]{fontenc}
\usepackage{cite}
\usepackage{balance}

\newtheorem{lemma}{Lemma}
\newtheorem{example}{Example}

\newcommand{\diag}{\mathop{\mathrm{diag}}}
\newcommand{\transpose}{\mathsf{T}}

\title{\LARGE \bf
Moment-based Kalman Filter: Nonlinear Kalman Filtering with Exact~Moment Propagation
}

\author{Yutaka Shimizu$^{1}$, Ashkan Jasour$^{3}$, Maani Ghaffari$^{4}$, and Shinpei Kato$^{1, 2}$
\thanks{Funding for M. Ghaffari was provided by NSF Award No. 2118818.}
\thanks{$^{1}$ Yutaka Shimizu and Shinpei Kato are with TIER IV, Inc., Jacom Building, 1-12-10 Kitashinagawa, Shinagawa-ku, Tokyo, 140-0001, Japan. {\tt\small \{shinpei.kato, yutaka.shimizu\}@tier4.jp}}%
\thanks{$^{2}$ Shinpei Kato are with the Graduate School of Information Science and Technology, The University of Tokyo, 7-3-1 Hongo, Bunkyo-ku, Tokyo, 113-0033, Japan.
}
\thanks{$^{3}$ Massachusetts Institute of Technology, Cambridge, MA 01239 USA {\tt\small jasour@mit.edu}. Now with NASA Jet Propulsion Lab,
California Institute of Technology.
}
\thanks{$^{4}$ Department of Naval Architecture and Marine Engineering, University of Michigan, Ann Arbor, MI 48109, USA. {\tt\small maanigj@umich.edu}
}}

\begin{document}

\maketitle
\thispagestyle{empty}
\pagestyle{empty}

\begin{abstract}
This paper develops a new nonlinear filter, called Moment-based Kalman Filter (MKF), using the exact moment propagation method. Existing state estimation methods use linearization techniques or sampling points to compute approximate values of moments. However, moment propagation of probability distributions of random variables through nonlinear process and measurement models play a key role in the development of state estimation and directly affects their performance. The proposed moment propagation procedure can compute exact moments for non-Gaussian as well as non-independent Gaussian random variables. Thus, MKF can propagate exact moments of uncertain state variables up to any desired order. MKF is derivative-free and does not require tuning parameters. Moreover, MKF has the same computation time complexity as the extended or unscented Kalman filters, i.e., EKF and UKF. The experimental evaluations show that MKF is the preferred filter in comparison to EKF and UKF and outperforms both filters in non-Gaussian noise regimes.

\end{abstract}

\section{INTRODUCTION}

The last few decades have witnessed a huge growth in nonlinear state estimation methods due to the advances in computer systems and sensors. This development enables us to safely navigate a robot even in a dynamic and uncertain environment, e.g., transporting a spaceship to a space station. Despite these advancements, we still have much room to improve state estimation methods, especially for nonlinear systems. Difficulties in nonlinear state estimation mainly arise from the complex probability distribution analysis. Generally, it is challenging to propagate accurate moments of uncertainties from a complex probability distribution, such as mean and covariance. Therefore, many traditional state estimation methods propagate approximated moments of uncertain states to estimate target values from noisy sensor inputs.

The majority of previous studies on state estimation are based on Kalman Filter (KF)~\cite{OrignalKalmanFilter,OriginalKalman2}. It aims to extract accurate information from noisy measurement data by minimizing the covariance of uncertain states. This linear optimal filter~\cite{KalmanFilterBook} can compute the exact mean and covariance of uncertain states when the system is linear. Even though KF is computationally efficient, its application is largely restricted because it can only handle a linear system. Therefore several studies have been made to broaden the scope of its application to enable nonlinear systems to use KF. 

The Extended Kalman Filter (EKF)~\cite{EKFBook,song1995asymptotic} is one of the most widely adopted nonlinear Kalman filters both in academics and industry\cite{EKFCILQG2021, EKFLQGMP2011, EKFVehiclePosition2020, EKFObjectTracking2017}. It usually uses first-order Taylor expansion to linearize nonlinear systems around the current estimated values, and moments of uncertainties are propagated through this linearized system. Hence the propagated moments of uncertain states have first-order accuracy, and they can achieve second-order accuracy when using the modified EKF~\cite{ModifiedEKF}.
Despite its wide applicability and computational advantage, it has several drawbacks. First, it often gets significant estimation errors when the system is highly nonlinear due to its limited range of approximation. In addition, EKF postulates that the system model has Jacobians or Hessians, which makes it difficult to apply to non-continuous systems. 

Unscented Kalman Filter (UKF) has been developed \cite{UKFPaper, UnscentedFilteringJulier, SquareRootUKF2002} to address these flaws of EKF. UKF uses a determined number of sampling points (sigma points) to approximate the probability distribution of uncertainty states and propagates each point through the nonlinear system. Then it computes approximated mean and covariance of transformed uncertain states using transformed sigma points~\cite{ANewExtensionJulier1997,TheScaledUnscentedJulier2002}. Since UKF can propagate moments of uncertain states up to third-order of accuracy in the presence of Gaussian distribution, it is also used in a wide range of applications\cite{UKFBeliefSpacePlanning2013, UKFStateEstimation2008, UKFTracking2018}. Although UKF is theoretically better than EKF, it still has some drawbacks. First, we have to tune several parameters to optimize the performance of UKF. Second, since UKF only uses a limited number of sampling points, its approximation accuracy decreases when the distribution has a large variance.

Apart from nonlinear Kalman filters, Particle Filter (PF)~\cite{doucet2001sequential,ProbablisticRobotics} is a well-known Monte Carlo-based approach for nonlinear state estimation~\cite{ristic2003beyond,ParticleFilterStateEstimation}. It approximates the state distribution with randomly sampled points (particles) and projects those particles into the nonlinear system function. Similar to UKF, PF also computes moments of uncertain states with these particles. Although the accuracy of approximated moments improves as the number of particles increases~\cite{ParticleFilterVSUKF}, the computation time also increases. Thus it hinders PF from being used in real-time applications in some domains.

In this paper, we improve the nonlinear Kalman filter by propagating exact moments of uncertain states. 
Unlike conventional approaches, the proposed Moment-based Kalman Filter (MKF) does not use system approximation or sampling methods for moment propagation. 
Instead, MKF propagates exact moments of state distributions by extending the recently proposed method \cite{ExactMomentPropagation,ExactMomentPropagationPlanning_2,ExactMomentPropagationPlanning}, which computes exact mixed-trigonometric-polynomial moments of uncertain states up to any desired order. 
The extended method of this paper can also compute exact mixed-trigonometric-polynomial moments of non-independent Gaussian random variables.
MKF has several major advantages over traditional state estimation methods. First, it uses exact moments of uncertain states to compute the estimation values with the same computational complexity as EKF and UKF.  
Moreover, MKF does not need to compute derivatives of the system, and also it does not have any parameters, which makes the proposed method a derivative-and-parameter-free filter.
These characteristics enable MKF to be used in a wide range of applications.

The contributions of this paper are as follows.
\begin{enumerate}
\item \textbf{Exact moment propagation procedure for estimation:}
The exact moment propagation method of MKF can compute exact mixed-trigonometric-polynomial moments of non-Gaussian random variables. Moreover, the provided extended exact moment propagation method can compute exact mixed-trigonometric-polynomial moments of Gaussian random variables even if they are not independent.

\item \textbf{A nonlinear derivative-and-parameter-free Kalman filter with exact moment propagation:} 
We incorporate the extended exact moment propagation into the Kalman filter to introduce a nonlinear derivative-and-parameter-free filter with the same calculation complexity as EKF and UKF for general state space problems.

\item  \textbf{Numerical experiments and simulations with real robot data:} 
Simulations with real robot data prove that the proposed algorithm has equivalent or even better performance than traditional approaches.
\end{enumerate}

The remaining of this paper is organized as follows. In Section~\ref{sec:extended_exact_moment_propagation}, we introduce the extended exact moment propagation method. Section~\ref{sec:moment_based_kalman_filter} delves into the proposed Moment-based Kalman filter. Numerical experiments and simulation results are presented in Section~\ref{sec:experiment}. Finally, Section~\ref{sec:conclusion} presents the conclusion and ideas for future work.

\section{Extended Exact Moment Propagation} \label{sec:extended_exact_moment_propagation}
In this section, we give details about the extended exact moment propagation method after briefly introducing the original exact moment propagation method~\cite{ExactMomentPropagation}.
\subsection{Exact Moment Propagation} \label{subsec:exact_moment_propagation}
The exact moment propagation method~\cite{ExactMomentPropagation} leverages the characteristic function~\cite{ProbabilityandStocasticsCinlar} of a random variable to compute the exact mixed-trigonometric-polynomial moments. 

\begin{lemma} \label{lemma:mixed-trigonometric-polynomial-moment}
    [\cite[Lemma~4]{ExactMomentPropagation}] Let $\theta$ be a random variable with characteristic function $\Phi_{\theta}(t)$, where $\Phi_{\theta}(t) = \mathbb{E}[e^{it\theta}]$. Given $(\alpha_1, \alpha_2, \alpha_3) \in \mathbb{N}^3$ where $\alpha = \sum_{i=1}^3 \alpha_i$, the following mixed-trigonometric-polynomial moments of order $\alpha$ of the form $\mathbb{E}[\theta^{\alpha_1}\cos^{\alpha_2}({\theta})\sin^{\alpha_3}({\theta})]$ can be computed as:
    
    \begin{equation}
    \begin{aligned}
    &\mathbb{E}[\theta^{\alpha_1}\cos^{\alpha_2}({\theta})\sin^{\alpha_3}({\theta})] =\\ 
    &\frac{1}{i^{\alpha_1 + \alpha_3} 2^{\alpha_2 + \alpha_3}} \sum_{(k_1, k_2)=(0, 0)}^{\alpha_2, \alpha_3} \begin{pmatrix} \alpha_2 \\ k_1 \end{pmatrix} \begin{pmatrix} \alpha_3 \\ k_2 \end{pmatrix}\\
    &\left. (-1)^{\alpha_3 - k_2} \frac{d^{\alpha_1}}{dt^{\alpha_1}}\Phi_{\theta}(t) \right|_{t=2(k_1 + k_2) - \alpha_2 - \alpha_3} .
    \end{aligned}
    \end{equation}
\end{lemma}
A more comprehensive description of this methodology and proof can be found in \cite{ExactMomentPropagation}.
In the following, we will extend Lemma 1 to compute the exact mixed-trigonometric-polynomial moments of non-independent Gaussian random variables. Let $\boldsymbol{x} \in \mathbb{R}^n$ be an $n$-dimensional random vector, and assume each element is independent. Given $(\alpha_i, c_i, s_i) \in \mathbb{N}^3$, the following mixed-trigonometric-polynomial moment can be computed as
\begin{align}
\label{eq:independent-vector-moment-calcuation}
    \nonumber \mathbb{E}\left[\prod_{i=1}^n x_i^{\alpha_i} \cos^{c_i}{(x_i)} \sin^{s_i}{(x_i)}\right] = \\
    \prod_{i=1}^n \mathbb{E}\left[x_i^{\alpha_i} \cos^{c_i}{(x_i)} \sin^{s_i}{(x_i)}\right],
\end{align}
where $x_i$ is $i$-th element in the random vector $\boldsymbol{x}$. With Lemma~\ref{lemma:mixed-trigonometric-polynomial-moment}, we can compute exact values for \eqref{eq:independent-vector-moment-calcuation}.
However, if each element in $\boldsymbol{x}$ is not independent, we cannot separately compute the nonlinear moments of $\boldsymbol{x}$. Therefore, we need to extend this method to enable exact moment propagation even when random variables are not independent.

\subsection{Variable Transformation} \label{sec:variable-transformation}
Before getting into the details about the extended exact moment propagation method, we introduce the variable transformation method for Gaussian random variables.
Let $\boldsymbol{x} \in \mathbb{R}^n$ be a n-dimensional non-independent Gaussian random vector, where $\boldsymbol{x} \sim \mathcal{N}(\boldsymbol{\mu}_x, \Sigma_x)$. Since $\Sigma_x$ is a real symmetric matrix, we can apply eigenvalue decomposition with an orthogonal matrix $T$ as
\begin{equation} \label{eq:eigen-decomposition-for-matrix}
\begin{aligned}
    T^{-1} \Sigma_x T = \Lambda =  \diag(\lambda_1, \lambda_2, \cdots, \lambda_n)
\end{aligned}
\end{equation}
where $\lambda_i$ is $i$-th eigenvalue of the covariance matrix $\Sigma_x$. Note that an orthogonal matrix $T$ satisfies $T^\transpose = T^{-1}$, $T^TT = I$ and $\mathrm{det} T = 1$ where $I$ is an Identity matrix and $T^\transpose$ is the transpose of the matrix $T$. Hence,
\begin{equation} \label{eq:orthgonal-det}
\begin{aligned}
    \mathrm{det}\Sigma_x &= \mathrm{det}T^{-1}\Lambda T = \mathrm{det} \Lambda \\
    \Sigma_x^{-1} &= \left(T \Lambda T^{-1}\right)^{-1} = T \Lambda^{-1} T^\transpose
\end{aligned}
\end{equation}
The probability density function of Gaussian random variable $\boldsymbol{x}$ can be described as:
\begin{equation}
    p_x(\boldsymbol{x}) = \frac{1}{\sqrt{(2\pi)^n \mathrm{det}\Sigma_x}} \exp{( -\frac{1}{2}(\boldsymbol{x} - \boldsymbol{\mu}_x)^\transpose \Sigma_x^{-1} (\boldsymbol{x} - \boldsymbol{\mu_x}) )}
\end{equation}
Linearly transform original variable $\boldsymbol{x}$, and define a new variable as $\boldsymbol{y} = T^{-1} \boldsymbol{x}$. Thus, we have $p_y(\boldsymbol{y}) =: \mathcal{N}(\boldsymbol{\mu}_y, \Sigma_y)$,
where $\boldsymbol{\mu}_y := T^\transpose\boldsymbol{\mu}_x$ and $\Sigma_y := \Lambda = \diag(\lambda_1, \lambda_2,..., \lambda_n)$. 
This indicates that $ \boldsymbol{y}$ also follows a Gaussian distribution, but each variable in $\boldsymbol{y}$ is independent since $\Sigma_y$ is a diagonal matrix, which suggests they are uncorrelated. Note that if each element in a Gaussian random vector is uncorrelated, they are also independent. 

\subsection{Extended Exact Moment Propagation}
Generally, it is hard to compute the nonlinear moments of non-independent random variables that come from arbitrary distributions. 
However, if we assume non-independent random variables follow a Gaussian distribution, we can compute exact mixed-trigonometric-polynomial moments by changing variables mentioned in Sec.\ref{sec:variable-transformation}. Given $(\alpha_i, c_i, s_i) \in \mathbb{N}^3$, mixed-trigonometric-polynomial moment of a Gaussian random vector $\boldsymbol{x}\in \mathbb{R}^n$ is
\begin{equation} \label{eq:original-mixed-trigonometric-polynomial-moment}
        \mathbb{E}\left[\prod_{i=1}^n x_i^{\alpha_i} \cos^{c_i}{(x_i)} \sin^{s_i}{(x_i)}\right] .
\end{equation}
Even after applying the linear transformation, the transformed moment is still mixed-trigonometric-polynomial. Given $(\beta_{k,i}, l_{k,i}, m_{k,i}) \in \mathbb{N}^3$, we can compute exact value for \eqref{eq:original-mixed-trigonometric-polynomial-moment} with transformed variable $\boldsymbol{y}$ as:
\begin{equation}
    \begin{aligned}
        &\mathbb{E}\left[\prod_{i=1}^n x_i^{\alpha_i} \cos^{c_i}{(x_i)} \sin^{s_i}{(x_i)}\right] \\
        & =\sum_{k} \mathbb{E}\left[ \prod_{i=1}^n \tilde{y}_i^{\beta_{k,i}} \cos^{l_{k,i}}{(\tilde{y}_{i})} \sin^{m_{k,i}}{(\tilde{y}_{i})} \right] \\
        & =\sum_{k} \prod_{i=1}^n \mathbb{E} \left[ \tilde{y}_i^{\beta_{k,i}} \cos^{l_{k,i}}{(\tilde{y}_{i})} \sin^{m_{k,i}}{(\tilde{y}_{i})} \right] \\
        & \left(\tilde{y}_i = g_k(T_{1,i}, T_{2,i}, \cdots , T_{n, i})y_i \right)
    \end{aligned}
\end{equation}
where $ g_k(T_{1,i}, T_{2,i}, \cdots , T_{n, i})$ is a constant value and $\tilde{y}_i$ is a scaled independent Gaussian random variable. Leveraging Lemma \ref{lemma:mixed-trigonometric-polynomial-moment}, we can compute the exact value for each $ \mathbb{E} \left[ \tilde{y}_i^{\beta_{k,i}} \cos^{l_{k,i}}{(\tilde{y}_{i})} \sin^{m_{k,i}}{(\tilde{y}_{i})} \right]$.

\begin{example}
Let $\boldsymbol{x} = \begin{pmatrix} x, \theta \end{pmatrix}^\transpose$ be a 2-dimensional non-independent Gaussian random variable vector, where \mbox{$\boldsymbol{x} \sim \mathcal{N}(\boldsymbol{\mu}, \Sigma)$}. The following mixed-trigonometric-polynomial moment can be computed by using the matrix $T$ and transformed independent Gaussian random variables $\boldsymbol{y} = (y_1, y_2)^\transpose \sim N \left(\begin{pmatrix} \overline{y}_1 \\ \overline{y}_2 \end{pmatrix}, \begin{pmatrix} \lambda_1 & 0 \\ 0 & \lambda_2 \end{pmatrix} \right)$ as:
\begin{equation} \label{eq:transformed-moment}
    \begin{aligned}
        &\mathbb{E}[x \cos{\theta}] = \iint \left(x \cos{\theta} \right) p_x(x, \theta) dx d\theta \\
        &= \iint (T_{11} y_1 + T_{12} y_2) \cos{(T_{21} y_1 + T_{22} y_2)} \\ 
        &\qquad \qquad \qquad \qquad \qquad \qquad p_y(y_1) p_y(y_2) dy_1 dy_2 \\
        &= \frac{T_{11}}{T_{21}^2T_{22}}\left\{ \mathbb{E}[l_1 \cos{l_1}] \mathbb{E}[\cos{l_2}] -  \mathbb{E}[l_1 \sin{l_1}] \mathbb{E}[\sin{l_2}]  \right\} \\
        &+ \frac{T_{12}}{T_{21}T_{22}^2}\left\{ \mathbb{E}[l_2 \cos{l_2}] \mathbb{E}[\cos{l_1}] -  \mathbb{E}[l_2 \sin{l_2}] \mathbb{E}[\sin{l_1}]  \right\},
    \end{aligned}
\end{equation}
    where $T_{ij}$ is $(i, j)$ element of matrix $T$, $l_1 \sim \mathcal{N}(T_{21} \overline{y}_1, T_{21}^2\lambda_1)$ and $l_2 \sim \mathcal{N}(T_{22}\overline{y}_2, T_{22}^2\lambda_2)$ are two independent Gaussian random variables. With Lemma \ref{lemma:mixed-trigonometric-polynomial-moment}, we can compute exact moment of $\mathbb{E}[x\cos{\theta}]$.
\end{example}

\section{Moment-Based Kalman Filter}  \label{sec:moment_based_kalman_filter}
The aim of this chapter is to improve the nonlinear Kalman filter by taking advantage of the exact moment propagation method discussed in Section \ref{sec:extended_exact_moment_propagation}.
Let $\boldsymbol{x}_k \in \mathbb{R}^n$ and $\boldsymbol{y}_k \in \mathbb{R}^m$ be a state vector and a measurement vector at time step $k$ respectively. With external disturbance $\boldsymbol{w}_k \in \mathbb{R}^d$ and measurement noise $\boldsymbol{v}_k \in \mathbb{R}^s$, a nonlinear discrete-time stochastic dynamics system can be modeled as:
\begin{equation} \label{eq:mkf-dynamics}
\begin{aligned}
    \boldsymbol{x}_{k+1} = f(\boldsymbol{x}_k, \boldsymbol{u}_k, \boldsymbol{w}_k), \quad 
    \boldsymbol{y}_{k+1} = h(\boldsymbol{x}_{k+1}, \boldsymbol{v}_{k+1}).
\end{aligned}
\end{equation}
where $\boldsymbol{u}_k$ is an input vector at time step $k$. In this paper, we assume dynamic and measurement models are described by mixed-trigonometric-polynomial functions. This is not a conservative assumption because these elementary functions can represent a wide range of robotic system models \cite{ExactMomentPropagationPlanning}. Note that external disturbance $\boldsymbol{w}_k$ and measurement noise $\boldsymbol{v}_k$ are independent of $\boldsymbol{x}_k$ and $\boldsymbol{y}_k$. 

Our method has four main stages: initial state distribution approximation, prediction, prior state distribution approximation, and update. 
Suppose we know the estimated mean $\hat{\boldsymbol{x}_k}$ and covariance $\hat{\Sigma}_{k}$ of previously estimated state $\boldsymbol{x}_k$, where $\boldsymbol{x}_k$ follows arbitrary distribution, and each element in $\boldsymbol{x}_{k}$ is not necessarily independent, which means non-diagonal elements in $\hat{\Sigma}_{k}$ can have non-zero values. In the initial state distribution approximation step, we approximate the previously estimated state distribution to a Gaussian distribution as:
\begin{equation} \label{eq:posterior-approximation}
    \boldsymbol{x}_{k}^{app} \sim \mathcal{N} \left( \hat{\boldsymbol{x}}_{k}, \hat{\Sigma}_{k} \right) .
\end{equation}
We use the mean and variance of the posterior distribution as the mean and variance of the approximated Gaussian distribution. 
In the prediction step, we predict the system transition from time step $k$ to time step $k+1$ and compute the predicted mean $\hat{\boldsymbol{x}}_{k+1}^p$ and covariance $\hat{\Sigma}_{k+1}^p$ as:
\begin{equation} \label{eq:prediction-step}
    \begin{aligned}
    \hat{\boldsymbol{x}}_{k+1}^p &= \mathbb{E}[\boldsymbol{x}_{k+1}]
    = \mathbb{E} \left[f(\boldsymbol{x}_k^{app}, \boldsymbol{u}_k, \boldsymbol{w}_k) \right], \\
    \hat{\Sigma}_{k+1}^p &= \mathbb{E} \left[(\boldsymbol{x}_{k+1} - \hat{\boldsymbol{x}}_{k+1}^p)(\boldsymbol{x}_{k+1} - \hat{\boldsymbol{x}}_{k+1}^p)^\transpose \right] \\
    &= \mathbb{E} \left[ \boldsymbol{x}_{k+1} \boldsymbol{x}_{k+1}^\transpose \right] - \left(\hat{\boldsymbol{x}}_{k+1}^p\right) \left(\hat{\boldsymbol{x}}_{k+1}^p\right)^\transpose \\
    &= \mathbb{E} \left[ f(\boldsymbol{x}_k^{app}, \boldsymbol{u}_k, \boldsymbol{w}_k) f(\boldsymbol{x}_k^{app}, \boldsymbol{u}_k, \boldsymbol{w}_k)^\transpose \right] \\
    & \qquad - \mathbb{E} \left[ f(\boldsymbol{x}_k^{app}, \boldsymbol{u}_k, \boldsymbol{w}_k) \right] \mathbb{E}\left[f(\boldsymbol{x}_k^{app}, \boldsymbol{u}_k, \boldsymbol{w}_k)\right]^\transpose.
    \end{aligned}
\end{equation}
Even though this is a nonlinear transformation of the non-independent Gaussian state $\boldsymbol{x}_k^{app}$, we can compute exact values for both $\hat{\boldsymbol{x}}_{k+1}^p$ and $\hat{\Sigma}_{k+1}^p$ with the extended moment propagation method as dynamics model $f(\boldsymbol{x}_k^{app}, \boldsymbol{u}_k, \boldsymbol{w}_k)$ are only composed of mixed-trigonometric-polynomial functions of $\boldsymbol{x}_k^{app}$, $\boldsymbol{u}_k$, and $\boldsymbol{w}_k$.
Due to the nonlinear transformation in the prediction step, $\boldsymbol{x}_{k+1}^{p}$ no longer follows Gaussian distribution. In order to compute the exact moment of uncertain states in the update step, we need to approximate the predicted state distribution to a Gaussian distribution.
Similar to the initial state distribution approximation, the approximated predicted state distribution $\boldsymbol{x}_{k+1}^{k, app}$ becomes:
\begin{equation} \label{eq:prior-approximation}
    \boldsymbol{x}_{k+1}^{p, app} \sim \mathcal{N} \left( \hat{\boldsymbol{x}}_{k+1}^p, \hat{\Sigma}_{k+1}^p \right).
\end{equation}
After the prior state distribution approximation, we can compute the predicted measurement mean and covariance. 
\begin{equation} \label{eq:predict-measurment-value}
\begin{aligned}
    \hat{\boldsymbol{y}}_{k+1} &= \mathbb{E}\left[h(\boldsymbol{x}_{k+1}^{p, app}, \boldsymbol{v}_{k+1})\right] \\
    \hat{\Sigma}_{k+1}^{yy} &= \mathbb{E} \left[ \left(\boldsymbol{y}_{k+1} - \hat{\boldsymbol{y}}_{k+1}\right)\left(\boldsymbol{y}_{k+1} - \hat{\boldsymbol{y}}_{k+1}\right)^\transpose \right] \\
    & = \mathbb{E} \left[ \boldsymbol{y}_{k+1} \boldsymbol{y}_{k+1}^\transpose \right] - \hat{\boldsymbol{y}}_{k+1}\hat{\boldsymbol{y}}_{k+1}^\transpose\\
    &= \mathbb{E} \left[ h(\boldsymbol{x}_{k+1}^{p, app},  \boldsymbol{v}_{k+1}) h(\boldsymbol{x}_{k+1}^{p, app},  \boldsymbol{v}_{k+1})^\transpose \right]  \\
    & \qquad - \mathbb{E}\left[h(\boldsymbol{x}_{k+1}^{p, app}, \boldsymbol{v}_{k+1})\right] \mathbb{E}\left[h(\boldsymbol{x}_{k+1}^{p, app}, \boldsymbol{v}_{k+1})\right]^\transpose .
\end{aligned}
\end{equation}
Similar to \eqref{eq:prediction-step}, we can compute exact values for both $\hat{\boldsymbol{y}}_{k+1}$ and $\hat{\Sigma}_{k+1}^{yy}$ by taking advantage of the extended moment propagation method as $\boldsymbol{x}_{k+1}^{p, app}$ is Gaussian random variables and $h(\boldsymbol{x}_{k+1}^{p, app}, \boldsymbol{v}_{k+1})$ is a mixed-trigonometric-polynomial function of $\boldsymbol{x}_{k+1}^{p, app}$ and $\boldsymbol{v}_{k+1}$.
Finally, we update the estimated state mean and covariance with measurement values $\boldsymbol{z}_{k+1}$ as:
\begin{equation}  \label{eq:update-step}
    \begin{aligned}
    \hat{\Sigma}_{k+1}^{xy} &= \mathbb{E} \left[ \left(\boldsymbol{x}_{k+1}^{p, app} - \hat{\boldsymbol{x}}_{k+1}^{p, app} \right)\left(\boldsymbol{y}_{k+1} - \hat{\boldsymbol{y}}_{k+1} \right)^\transpose \right] \\
    &= \mathbb{E} \left[ \boldsymbol{x}_{k+1}^{p, app} h(\boldsymbol{x}_{k+1}^{p, app}, \boldsymbol{v}_{k+1})^\transpose \right] - \hat{\boldsymbol{x}}_{k+1}^{p, app}\hat{\boldsymbol{y}}_{k+1}^\transpose \\
    K_{k+1} &= \hat{\Sigma}_{k+1}^{xy} \left( \hat{\Sigma}_{k+1}^{yy} \right)^{-1} \\
    \hat{\boldsymbol{x}}_{k+1} &= \hat{\boldsymbol{x}}_{k+1}^p + K_{k+1}(\boldsymbol{z}_{k+1} - \hat{\boldsymbol{y}}_{k+1}) \\
    \hat{\Sigma}_{k+1} &= \hat{\Sigma}_{k+1}^p - K_{k+1} \hat{\Sigma}_{k+1}^{yy} K_{k+1}^\transpose
    \end{aligned}
\end{equation}
where $K_{k+1}$ is Kalman Gain. Since $h(\boldsymbol{x}_{k+1}^{p, app}, \boldsymbol{v}_{k+1})$ is a mixed-trigonometric-polynomial function, $\mathbb{E} \left[ \boldsymbol{x}_{k+1}^{p, app} h(\boldsymbol{x}_{k+1}^{p, app}, \boldsymbol{v}_{k+1}) \right]$ is also a mixed-trigonometric-polynomial moment of $\boldsymbol{x}_{k+1}^{p, app}$ and $\boldsymbol{v}_{k+1}$, and thus we can compute the exact value of $\hat{\Sigma}_{k+1}^{xy}$. The overall procedure is summarized in Algorithm~\ref{alg:mkf}.

\begin{algorithm}[t]
    \footnotesize
    \SetAlgoLined
    \SetNoFillComment
    \SetKwData{Left}{left}\SetKwData{This}{this}\SetKwData{Up}{up}
    \SetKwFunction{Union}{Union}\SetKwFunction{FindCompress}{FindCompress}
    \SetKwInOut{Input}{input}\SetKwInOut{Output}{output}
    \Input{Current estimated mean $\hat{\boldsymbol{x}}_k$ \\
           Current estimated variance $\hat{\Sigma}_k$ \\
           Measurement value $\boldsymbol{z}_{k+1}$}
    \Output{Next estimated mean $\hat{\boldsymbol{x}}_{k+1}$ \\
            Next estimated variance $\hat{\Sigma}_{k+1}$}
    $\hat{\boldsymbol{x}}_k^{app}$, $\hat{\Sigma}_k^{app}$ $\leftarrow$ Initial Approximation($\hat{\boldsymbol{x}}_k$, $\hat{\Sigma}_k$) \Comment{\eqref{eq:posterior-approximation}}\\
    $\hat{\boldsymbol{x}}_{k+1}^{p}$, $\hat{\Sigma}_{k+1}^{p}$ $\leftarrow$ Predict($\hat{\boldsymbol{x}}_k^{app}$, $\hat{\Sigma}_k^{app}$) \Comment{\eqref{eq:prediction-step}}\\
    $\hat{\boldsymbol{x}}_{k+1}^{p, app}$, $\hat{\Sigma}_{k+1}^{p, app}$ $\leftarrow$ Prior Approximation($\hat{\boldsymbol{x}}_{k+1}^{app}$, $\hat{\Sigma}_{k+1}^{app}$) \Comment{\eqref{eq:prior-approximation}}\\
    $\hat{\boldsymbol{x}}_{k+1}$, $\hat{\Sigma}_{k+1}$ $\leftarrow$ Update($\hat{\boldsymbol{x}}_{k+1}^{p, app}$, $\hat{\Sigma}_{k+1}^{p, app}$, $\boldsymbol{z}_{k+1}$) \Comment{\eqref{eq:predict-measurment-value}-\eqref{eq:update-step}}\\
    \Return $\hat{\boldsymbol{x}}_{k+1}$, $\hat{\Sigma}_{k+1}$
    \caption{Moment-Based Kalman Filter}
    \label{alg:mkf}
\end{algorithm}

MKF has two major advantages over EKF and UKF. First, MKF does not require Jacobians for the calculation. This is a great advantage over EKF because we can apply MKF for non-continuous systems. Next, MKF does not have any tuning parameters; thus, it does not need to find the optimal parameters to get the best performance. UKF, on the contrary, has several parameters, and we need to tune these parameters to improve performance. These two points make the proposed filter a derivative-and-parameter-free filter. 
In addition to these two advantages, MKF has the same computation complexity as EKF and UKF for general state space problems\cite{SquareRootUKF2002}. Since the most computationally heavy part is Eigen decompositions which are required in computing the exact moment of non-independent state variables in \eqref{eq:prediction-step}-\eqref{eq:update-step}, the overall computational complexity of MKF becomes $\mathcal{O}(n^3)$ where $n$ is the dimension of the state vector $\boldsymbol{x}_k$. The comparison of EKF, UKF, and MKF is described in Table~\ref{table:kf-comparision}.

\begin{table} 
  \begin{center}
    \caption{Comparision of nonlinear Kalman Filters.}
    \label{table:kf-comparision}
    \begin{tabular}{|c|c|c|c|}
      \hline
      & EKF & UKF & MKF \\
      \hline
      Jacobian & Required & No & No\\
      \hline
      \begin{tabular}{c} Number of\\ parameters\end{tabular} & 0 & several & 0 \\
      \hline
      \begin{tabular}{c} Computational\\ complexity\end{tabular} &  $\mathcal{O}(n^3)$  &  $\mathcal{O}(n^3)$ &  $\mathcal{O}(n^3)$  \\
      \hline
      Model forms & \begin{tabular}{c} Differentiable \\ (Continuous) \end{tabular} & any & \begin{tabular}{c} mixed \\trigonometric \\polynomial \end{tabular}\\
      \hline
    \end{tabular}
  \end{center}
\end{table}

\section{Experimental Results}  \label{sec:experiment}
In this section, we evaluate the extended exact moment propagation method and MKF through numerical experiments and simulations with real robot data. In each experiment, we compare the proposed method with other existing methods to prove the effectiveness of the proposed approach. All of the experiments are tested on a desktop computer with Intel i7 4.2 GHz processors and 32 GB RAM, and test codes are written in C++. Since formulating equations to compute the high order of moments of uncertainty states is easily subject to human errors, we use TreeRing to automatically compute formulations of moment propagation equations\cite{TreeRing1}\cite{TreeRing2}.
Our software is available for download\footnote{\label{footnote1}\url{https://github.com/purewater0901/MKF.git}}.

\subsection{Numerical experiments for extended exact moment propagation}
The objective of numerical experiments is to validate the extended exact moment propagation (Extended EMP) method. To achieve this goal, we compare the proposed method with Linear transformation, Unscented Transform (UT), and the original exact moment propagation method (Original EMP). 
When using the original EMP, we compute the moments of each random variable separately, as shown in \eqref{eq:independent-vector-moment-calcuation} and ignore the correlation between random variables since it cannot handle non-independent random variables.
We verify true values by Monte Carlo simulation with $10^8$ samples. 

Table~\ref{tab:numerical_simulation_1} shows the computation results of the mixed-trigonometric-polynomial moments of independent non-Gaussian distributions. 
When each element is only composed of linear functions, e.g., $\mathbb{E}[x\theta]$, all methodologies can compute exact moments. However, when the function becomes highly nonlinear, Linear and UT cannot provide exact values. Since both the original EMP and the proposed method can compute moments of independent non-Gaussian distributions, they get true values verified by Monte Carlo simulation.
Table~\ref{tab:numerical_simulation_2} presents several nonlinear moment propagation results of a non-independent two-dimensional Gaussian random variable. 
It indicates that as the function becomes highly nonlinear or two random variables $x$ and $\theta$ have stronger correlations, Linear propagation and UT have large errors. In addition, the original EMP also has large deviations from true values obtained by Monte Carlo simulation because it cannot compute exact mixed-trigonometric-polynomial moments of non-independent Gaussian distributions. 
On the contrary, the proposed method obtains the true values verified by Monte Carlo simulation in all cases, as it can handle non-independent Gaussian distributions.
Note that UT can compute the exact values for $\mathbb{E}[xy]$ since it can calculate the exact moment up to the third order of nonlinearity. 
Furthermore, Table~\ref{tab:numerical_simulation_3} provides the results of a three-dimensional Gaussian moment propagation. Similar to the previous result, the proposed algorithm only gives the same values as the one obtained by Monte Carlo simulation. Overall the results prove that the proposed method can only compute exact mixed-trigonometric-polynomial moments of non-independent Gaussian random variables.

\begin{table}
  \begin{center}
    \caption{$x \sim Exponential(1.0)$, ~ $\theta \sim Uniform(-\frac{\pi}{3}, \frac{\pi}{6})$ }
    \label{tab:numerical_simulation_1}
    \resizebox{\columnwidth}{!}{
    \begin{tabular}{|c|c|c|c|} 
      \hline
      \diagbox{Methodology}{Value} & $\mathbb{E}[x\theta]$ & $\mathbb{E}[x\cos{\theta}]$ & $\mathbb{E}[x\cos{\theta}\sin{\theta}]$ \\
      \hline
      Linear & -0.262 & 0.966 & -0.25 \\
      \hline
      UT & -0.262 & 0.875 & -0.178 \\
      \hline
      Original EMP & -0.262 & 0.870 & -0.159 \\
      \hline
      Extended EMP (Proposed) & -0.262 & 0.870 & -0.159 \\
      \hline
      Monte Carlo $(N=10^8)$ & -0.262 & 0.870 & -0.159\\
      \hline
    \end{tabular}
    }
  \end{center}
\end{table}

\begin{table}
  \begin{center}
    \caption{$\begin{pmatrix}x \\ \theta \end{pmatrix} \sim \mathcal{N} \left( \begin{pmatrix} 10.0 \\ \frac{\pi}{3}\end{pmatrix}, \begin{pmatrix} 5.0 & 1.5 \\ 1.5 & \frac{\pi}{6} \\ \end{pmatrix} \right)$ }
    \label{tab:numerical_simulation_2}
    \resizebox{\columnwidth}{!}{
    \begin{tabular}{|c|c|c|c|} 
      \hline
      \diagbox{Methodology}{Value} & $\mathbb{E}[x\theta]$ & $\mathbb{E}[x\cos{\theta}]$ & $\mathbb{E}[x\cos{\theta}\sin{\theta}]$ \\
      \hline
      Linear & 10.47 & 5.000 & 0.433\\
      \hline
      UT & 11.97 & 3.028 & 2.008\\
      \hline
      Original EMP & 10.47 & 3.848 & 1.520 \\
      \hline
      Extended EMP (Proposed) & 11.97 & 2.848 & 1.256 \\
      \hline
      Monte Carlo $(N=10^8)$ & 11.97 &  2.848 & 1.256 \\
      \hline
    \end{tabular}
    }
  \end{center}
\end{table}

\begin{table}
  \begin{center}
    \caption{$\begin{pmatrix}x \\ y \\ \theta \end{pmatrix} \sim \mathcal{N} \left( \begin{pmatrix} 10.0 \\ 5.0 \\ \frac{\pi}{3}\end{pmatrix}, \begin{pmatrix} 3.0 & 0.5 & 0.5 \\ 0.5 & 2.0 & 0.3 \\ 0.5 & 0.3 & \frac{\pi}{10} \\ \end{pmatrix} \right) $}
    \label{tab:numerical_simulation_3}
    \begin{tabular}{|c|c|c|} 
      \hline
      \diagbox{Methodology}{Value} & \textbf{$\mathbb{E}[xy\sin{\theta}]$} & \textbf{$\mathbb{E}[x^2y\cos{\theta}]$} \\
      \hline
      Linear & 43.30 & 257.5 \\
      \hline
      UT & 39.94 & 157.3 \\
      \hline
      Original EMP & 37.01 & 220.1 \\
      \hline
      Extended EMP (Proposed) & 39.62 & 162.3 \\
      \hline
      Monte Carlo $(N=10^8)$ & 39.62 & 162.3\\
      \hline
    \end{tabular}
  \end{center}
\end{table}

\subsection{Simulation with real robot data}
Next, we evaluate MKF with real robot data using UTIAS Multi-Robot Cooperative Localization and Mapping Dataset\cite{SimulationOpenDataset}. This dataset includes the ground truth 2D position $x$, $y$, and yaw angle $\theta$ of the robot, odometry data recording input values to the robot, and noisy measurement data. The ground truth robot state $\boldsymbol{x} = (x, y, \theta)$ data are gathered through their 10-camera Vicon motion capture system with accuracy on the order of $0.001[m]$. Odometry data contains forward velocity command $v$ and angular velocity command $u$ at each time. Since they use two-wheel differential drive robots to create the dataset, the discrete dynamical model of the robot can be described as:
\begin{equation}
\begin{aligned}
    x(k+1) &= x_k + (v(k) + \omega_v(k)) \cos{\theta(k)}dt \\
    y(k+1) &= x_k + (v(k) + \omega_v(k)) \sin{\theta(k)}dt \\
    \theta(k+1) &= \theta(k) + (u(k)+\omega_u(k))dt,
\end{aligned}
\end{equation}
where $dt$ is the sampling time, and $\omega_v$ and $\omega_u$ are external disturbances. 

Measurement data in this dataset contain range $r$ and bearing angle $\varphi$ to the landmarks. Bearing angle $\varphi$ can be defined as
\begin{equation}
    \varphi(k) = \arctan\left({\frac{y_l^i - y(k)}{x_l^i - x(k)}}\right) - \theta(k)
\end{equation}
where $x_l^i$ and $y_l^i$ are 2D positions of the $i$-th landmark. 
In this experiment, we assume measurement values of range $\hat{r}$ and bearing angle $\hat{\varphi}$ have multiplicative and additive noise, respectively, e.g., $\hat{r} = r \cdot v_r$ and $\hat{\varphi} = \varphi +v_{\varphi}$.
In order to work with mixed-trigonometric-polynomial measurement functions, we define the measurement model as follows:
\begin{equation} \label{eq:experiment-mtp-measurement-model}
\begin{aligned}
    y_1(k) = \hat{r}(k) \cos\left(\hat{\varphi}(k)\right) , \quad
    y_2(k) = \hat{r}(k) \sin\left(\hat{\varphi}(k)\right) 
\end{aligned}
\end{equation}
By expanding the \eqref{eq:experiment-mtp-measurement-model}, the measurement functions become
\begin{equation}
\begin{aligned}
    y_1(k) &= r(k) v_r(k) \cos\left({\varphi(k) + v_{\varphi}(k)}\right) \\
    &= h_a(k) v_r(k) \cos\left({v_{\varphi}(k)}\right) - h_b(k) v_r(k) \sin\left({v_{\varphi}(k)}\right), \\
    y_2(k) &= r(k) v_r(k) \sin\left({\varphi(k) + v_{\varphi}(k)}\right) \\
    &= h_b(k) v_r(k) \cos\left({v_{\varphi}(k)}\right) + h_a(k) v_r(k) \sin\left({v_{\varphi}(k)}\right),
\end{aligned}
\end{equation}
where 
\begin{equation}
\begin{aligned}
    h_a(k) &= \left(x_l^i - x(k)\right) \cos\left({\theta(k)}\right) + \left(y_l^i - y(k)\right) \sin\left({\theta(k)}\right). \\
    h_b(k) &= \left(y_l^i - y(k)\right) \cos\left({\theta(k)}\right) - \left(x_l^i - x(k)\right) \sin\left({\theta(k)}\right).
\end{aligned}
\end{equation}

The initial uncertain state $\boldsymbol{x}(0) = (x(0), y(0), \theta(0))$ is sampled from the following Gaussian distribution 
\begin{equation}
    \boldsymbol{x}(0) \sim \mathcal{N}\left( \begin{pmatrix} 3.573 \\ -3.333 \\ 2.341 \end{pmatrix}, 
    \begin{pmatrix} 0.01^2 & 0.0 & 0.0 & \\ 0.0 & 0.01^2 & 0.0 \\ 0.0 & 0.0 & 0.01^2 \end{pmatrix} \right)
\end{equation}

In this simulation, we set two environments where external disturbance and measurement noise follow different distributions. In both environments, we compare MKF with EKF and UKF. Note that the parameters of UKF are determined from \cite{UKFParameter2012} as they make UKF more stable. 

\subsubsection{Gaussian disturbance and noise}
In this setting, we assume that original data contain the following Gaussian external disturbance and Gaussian measurement noise. 
\begin{equation}
\begin{aligned}
 \omega_v(k) &\sim \mathcal{N}(0.0, 0.01), \quad \omega_{u}(k) \sim \mathcal{N}(0.0, 1.0) \\
 v_r(k) &\sim \mathcal{N}(1.0, 0.01), \quad v_{\varphi}(k) \sim \mathcal{N}(0.0, 0.0007).
\end{aligned}
\end{equation}

\subsubsection{Gaussian disturbance and Non-Gaussian noise}
In this setting, we assume measurement noise is generated from non-Gaussian distributions. Since the original measurement data already contains noise, we recalculate the measurement values from the ground truth data. After that, we collapse the true measurement values with the following non-Gaussian noise.
\begin{equation}
\begin{aligned}
 \omega_v(k) &\sim \mathcal{N}(0.0, 0.01), \quad \omega_{u}(k) \sim \mathcal{N}(0.0, 1.0) \\
 v_r(k) &\sim Exponential(1.0) \\
 v_{\varphi}(k) &\sim Uniform(-v_{\varphi, \text{limit}}, v_{\varphi, \text{limit}})
\end{aligned}
\end{equation}

\begin{figure}[t] 
    \centering
    \subfloat[]{\includegraphics[scale=0.20]{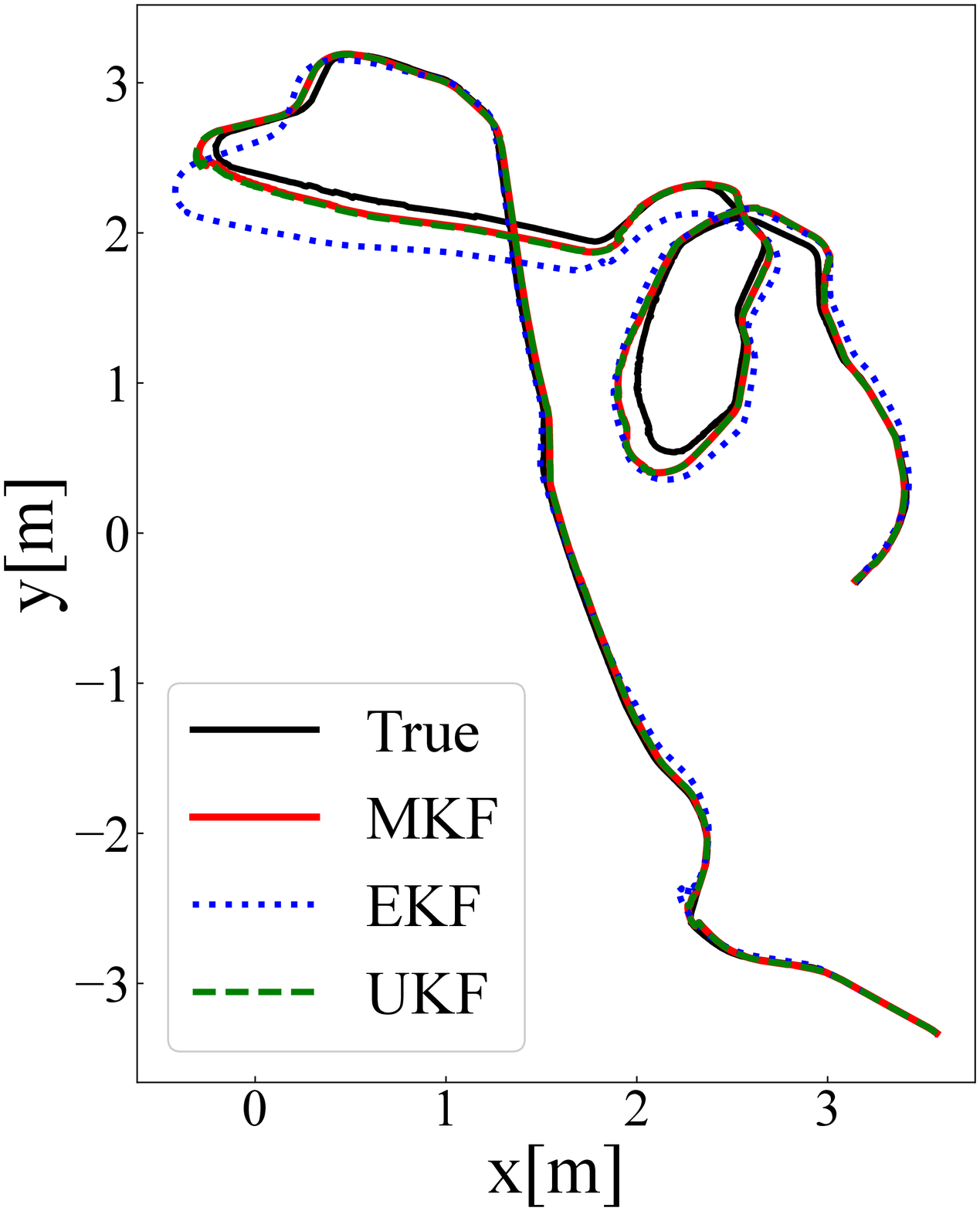}} 
    \quad
    \subfloat[]{\includegraphics[scale=0.20]{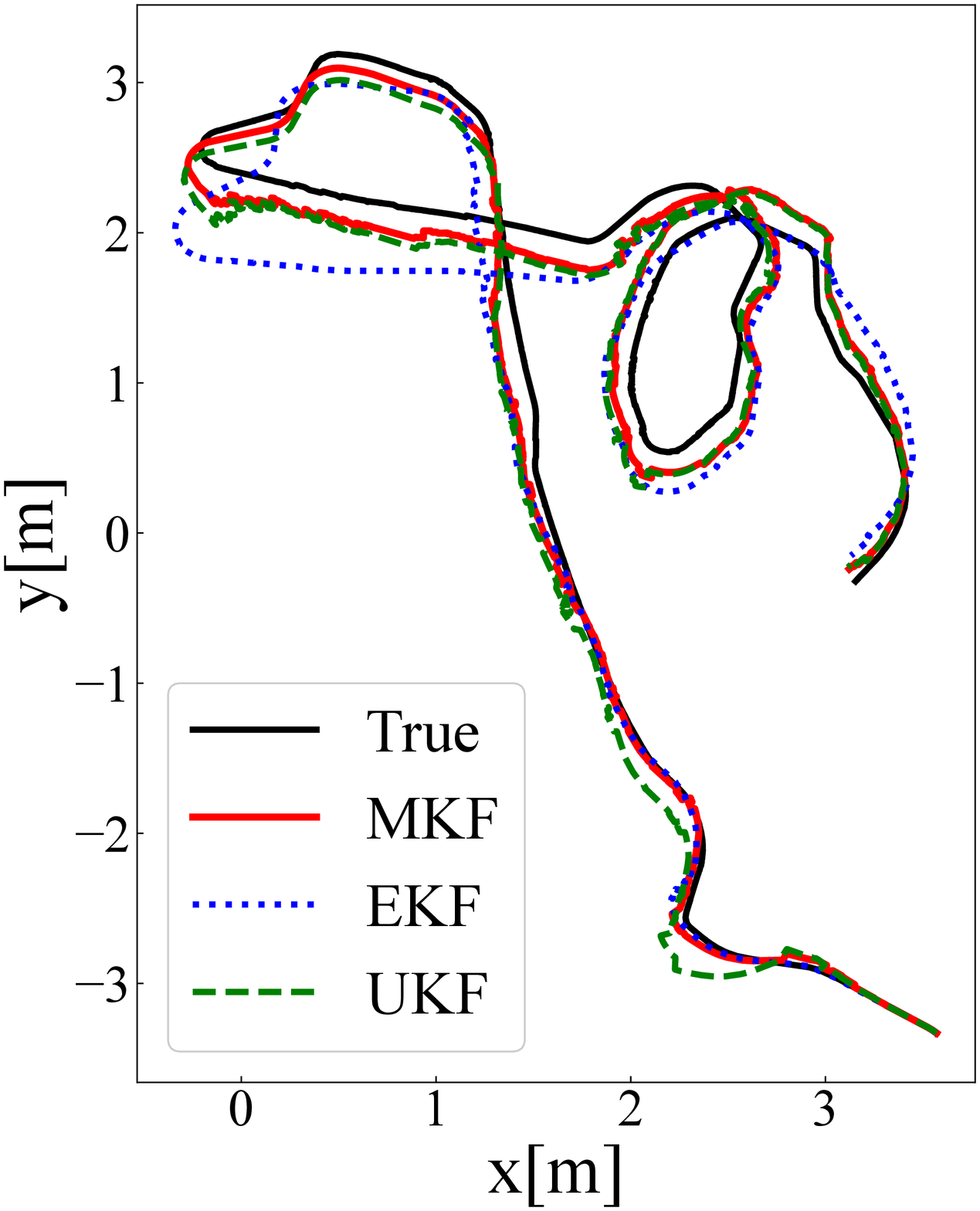}} 
    \caption{Simulation results with Gaussian noise (a) and non-Gaussian noise (b). Both pictures depict true trajectory and estimated trajectories by EKF, UKF, and MKF.}
    \label{fig:kalman-filter-simulation-real-data-result}
\end{figure}

\begin{figure}[t] 
    \centering
    \subfloat[]{\includegraphics[scale=0.20]{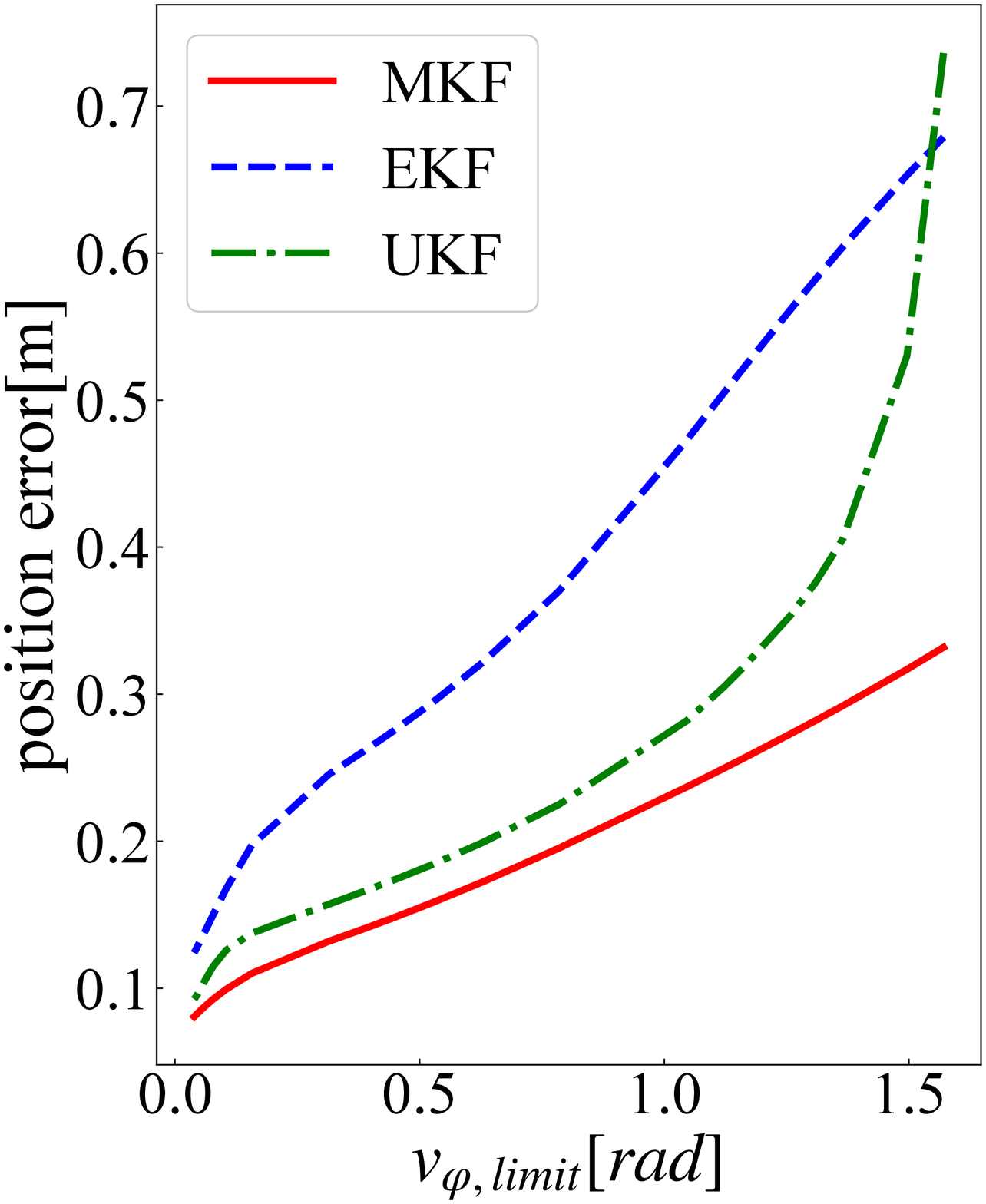}} 
    \quad
    \subfloat[]{\includegraphics[scale=0.20]{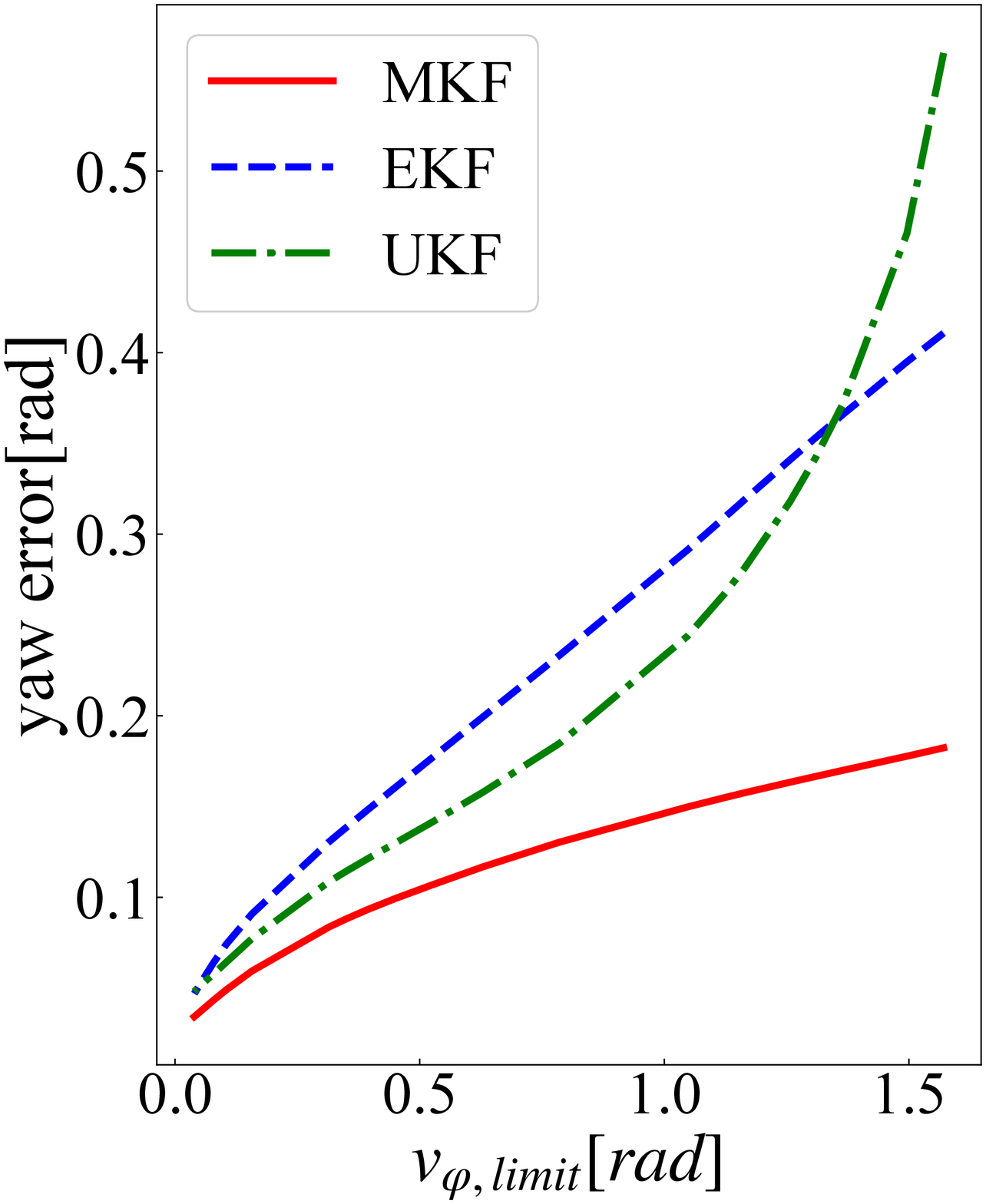}} 
    \caption{Simulation results when increasing the measurement noise of the bearing angle. (a) and (b) shows position and yaw errors, respectively.}
    \label{fig:kalman-filter-simulation-progression}
\end{figure}

Both simulation results are presented in Table~\ref{tab:real-data-simulation-mean-error} and Fig.~\ref{fig:kalman-filter-simulation-real-data-result}. Note that in the non-Gaussian setting, we set the limitation of measurement noise of bearing angle as $v_{\varphi, \text{limit}} = \frac{\pi}{12}$.
Table~\ref{tab:real-data-simulation-mean-error} shows the mean position and yaw errors over the simulation time. Note that position error is measured by Euclidean distance of the estimated point and ground truth point, and yaw errors are measured by absolute error. 
We can find from the results that MKF has the smallest deviations from the ground truth values under the Gaussian and non-Gaussian measurement values. For Gaussian conditions, UKF also performs as well as MKF. This is because UKF can compute moments accurately until the third order of nonlinearity in the presence of Gaussian noise, allowing UKF to give highly accurate estimations. However, when the measurement noise follows non-Gaussian distributions, MKF largely outperforms UKF. The large errors of estimation stem from the characteristic of UKF that it can only guarantee the second order of nonlinearity in the presence of non-Gaussian noise. Hence, UKF has large approximation errors when propagating moments through nonlinear measurement equations. Since EKF has only first-order accuracy, it has the largest deviations in these nonlinear Kalman filters. 
In addition, we increase the measurement noise of the bearing angle $v_{\varphi, \text{limit}}$ to see how each method's position and yaw errors are increasing in the non-Gaussian environment. 
Fig.~\ref{fig:kalman-filter-simulation-progression} shows that the position and yaw errors of UKF get much larger as the measurement noise of the bearing angle increases. At the same time, MKF does not have a significant increase in position and yaw errors. This is because when the noise variance becomes large, the Unscented Transformation gets large approximation errors of the computed moments, and its error easily diverges. MKF, on the other hand, can propagate exact moments of uncertain states even when the noise has a large variance thanks to the exact moment propagation method; thus, it has robust results for any magnitude of the variance.

\begin{table}[t]
  \begin{center}
    \caption{Mean error of estimated values with Gaussian measurement noise and non-Gaussian measurement noise}
    \label{tab:real-data-simulation-mean-error}
    \begin{tabular}{|c||c|c|c||c|c|c|}
      \hline
      & \multicolumn{3}{c||}{Gaussian Noise} & \multicolumn{3}{c|}{Non-Gaussian Noise} \\
      \hline
      & EKF & UKF & MKF & EKF & UKF & MKF \\
      \hline
      Position[m] & 0.12 & 0.058 & 0.057 & 0.20 & 0.15 & 0.11 \\
      \hline
      Yaw[rad] & 0.060 & 0.037  & 0.037 & 0.12 & 0.11  & 0.08 \\
      \hline
    \end{tabular}
  \end{center}
  \vspace{-3mm}
\end{table}

\section{Conclusion}  \label{sec:conclusion}
In this paper, we proposed the extended exact moment-based uncertainty propagation algorithm and Moment-based Kalman filter. The proposed uncertainty propagation allows us to compute exact mixed-trigonometric-polynomial moments of non-independent Gaussian random variables. Combining the proposed extended moment propagation method with the Kalman filter, MKF can propagate exact moments of uncertain states without introducing parameters and calculating derivatives of the system. Numerical experiments and simulations with real robot data show that the proposed approaches have the same or better results than compared baselines with the same computation complexity order.

Although the MKF has successfully demonstrated its superior performance to traditional nonlinear Kalman filters, it has certain limitations in terms of handling higher-order moments. Due to Gaussian distribution approximations, it only propagates first and second-order moments, which reduces higher-order moments after the nonlinear transformations. To overcome this limitation, we should consider extending MKF to handle higher-order moments in the future. Another possible extension of MKF is to apply it to the belief space planning problem. Finally, an attractive research direction is an extension of the moment-based approach to symmetry-preserving estimation and control methods~\cite{bonnabel2008symmetry,barrau2017invariant,hartley2020contact,van2020equivariant,mahony2021equivariant,9981282,9993143,ghaffari2022progress}.




{
\balance
\bibliographystyle{IEEEtran}
\bibliography{strings-full,ieee-full,reference}
}

\end{document}